\documentclass{article} % For LaTeX2e
\usepackage{iclr2023_conference,times}

\usepackage{amsmath,amsfonts,bm}

\def\eqref#1{equation~\ref{#1}}
\def\1{\bm{1}}

\DeclareMathAlphabet{\mathsfit}{\encodingdefault}{\sfdefault}{m}{sl}
\SetMathAlphabet{\mathsfit}{bold}{\encodingdefault}{\sfdefault}{bx}{n}

\usepackage{hyperref}
\usepackage{url}
\usepackage{comment}
\usepackage{graphicx}
\usepackage{pbox}

\title{Enabling Calibration In The Zero-Shot Inference of Large Vision-Language Models}

\author{Will Levine$^\dagger$, Benjamin Pikus$^\dagger$, Pranav Raja \& Fernando Amat Gil \\ 
Scale AI \\
\texttt{\{will.levine,ben.pikus,pranav.raja,fernando.gil\}@scale.com} \\
}

\iclrfinalcopy % Uncomment for camera-ready version, but NOT for submission.
\begin{document}

\maketitle

\begin{abstract}
Calibration of deep learning models is crucial to their trustworthiness and safe usage, and as such, has been extensively studied in supervised classification models, with methods crafted to decrease miscalibration. However, there has yet to be a comprehensive study of the calibration of vision-language models that are used for zero-shot inference, like CLIP. We measure calibration across relevant variables like prompt, dataset, and architecture, and find that zero-shot inference with CLIP is miscalibrated. Furthermore, we propose a modified version of temperature scaling that is aligned with the common use cases of CLIP as a zero-shot inference model, and show that a single learned temperature generalizes for each specific CLIP model (defined by a chosen pre-training dataset and architecture) across inference dataset and prompt choice.
\end{abstract}

{\let\thefootnote\relax\footnotetext{$^{\dagger} $ denotes equal contribution}}

\section{Introduction}

Interpretability is one of the main hurdles in the trust, safety, and reliability of deep learning models. One specific area of concern is the miscalibration of these models, where the confidences of the model predictions do not reflect the probabilities of being correct. There exists many studies on the calibration and corresponding interpretability of classification models \citep{guo2017calibration, kull2019beyond, rajendran2019accurate} that are trained and tested on in a traditional fashion - a given dataset of one modality (like images) is split into a train, validation, and test set, with a known, fixed number of classes. The model is trained on the train set, tuned with the validation set, and evaluated on the test set. However, the use of vision-language models for zero-shot inference, like CLIP \citep{radford2021learning}, is becoming increasingly popular. In this setting, the dataset is multimodal, and the inference paradigm allows for zero-shot inference, where the class being predicted was not explicitly defined as a class of interest during training. 

There has yet to be either an extensive study of the calibration of CLIP as a zero-shot inference model or applications of calibration methods to CLIP's zero-shot inference setting. We therefore propose ``Zero-Shot-Enabled Temperature Scaling," a method based on Temperature Scaling (TS) that enables zero-shot inference for models like CLIP. Our main \textbf{contributions} are

\begin{itemize}
    \item An extensive analytical study of the calibration of CLIP stratified by architecture, dataset (pre-training and inference), and input prompt. There exists literature that briefly mentions the calibration of CLIP \citep{minderer2021revisiting}, but not one that studies the calibration across any of the previously mentioned experimental variables.
    \item The novel application of Temperature Scaling to CLIP in a way that preserves the ability for zero-shot inference with an exposition on its robustness to changes in inference dataset and prompt. We note that the only modification to inference is to perform Temperature Scaling on the text-image similarity with temperature $T$. We show that this parameter varies solely with changes in underlying architecture and pre-training dataset (identical to the axes of variation allowed in the parameters of CLIP), and thus can be used at inference time for any arbitrary set of prompts or inference datasets. Therefore, to perform inference on a given dataset of interest requires no training, tuning, or calibration - meaning our method matches the zero-shot inference paradigm as used in CLIP. %If this paper is accepted, we will make these $T$ parameters per architecture/pre-training dataset pair publicly available, although we do note that users can train this temperature $T$ themselves in a way that is described in Section \ref{calibrate_labels}.

\end{itemize}

\section{Preliminaries}
\subsection{Problem Setup}
\label{problem_setup}
Let $X: \Omega \to \mathcal{X} \subset \mathbb{R}^D$ be the input (random) variable and let $Y: \Omega \to \{1,2,...,C-1,C\} \subset \mathbb{N}$ (where $C$ is the number of output classes) be the response (random) variable. Typically $X$ has some information about $Y$ and we'd like to make inferences about $Y$ given $X$.
A common situation is trying to compute $\underset{c}{\mathrm{argmax}}\hspace{0.5mm}\mathbb{E}(Y = c | X = x)$ with a model $\hat{f}: \mathbb{R}^d \to [0, 1]^C$. That is, $\hat{f}$ predicts the most likely class on inference example $x_i$ among the output classes as $\underset{c}{\mathrm{argmax}}\hat{f}_c(x_i)$. Typically, $f$ uses an intermediate logit function $L : \mathbb{R}^D \to \mathbb{R}^C$. That is, the logit function returns a real number per class. For a given class index $c$, the logit corresponding to that class $L_c(x_i)$ ideally increases as the resemblance increases between the inference example $x_i$ and the training inputs of class $c$ (formally $\{x_j \hspace{1mm}\text{s.t.}\hspace{1mm}y_j = c\}_{j=1}^N$).

\subsection{CLIP}

Introduced in \citet{radford2021learning}, CLIP is trained to align image/text pairs. This enables zero-shot inference, where the output classes of $Y$ are chosen at inference time based on natural language prompts. CLIP accomplishes this by calculating its logit function as the cosine similarity between the embedding of a given input image $x_i$ - which we denote as $E_{\text{im}}(x_i)$ - and the language embeddings of the natural language corresponding to the output classes. We denote the language embedding of the natural language corresponding to a given output class $y_c$ as $E_{\text{lang}}(y_c)$. More formally, the logit function of CLIP is\footnote{The $100$ in the logit function is a standard scalar temperature used as a multiplier for the CLIP image-similarity logit, as seen in \url{https://github.com/openai/CLIP/} } $$L^{\text{CLIP}}_c(x_i) = 100 * \dfrac{E_{\text{im}}(x_i) \cdot E_{\text{lang}}(y_c)}{|E_{\text{im}}(x_i)||E_{\text{lang}}(y_c)|}$$

For CLIP and non-CLIP models, the softmax function is typically used to convert these logits into class probabilities. That is $$\hat{f}_c(x_i) = \dfrac{e^{L^{\text{CLIP}}_c(x_i)}}{\sum_{j=1}^Ce^{L^{\text{CLIP}}_j(x_i)}}$$

\subsection{Calibration} 

Ideally, we would like the confidence estimate $\hat{p}(x_i, \hat{f}) = \underset{c}{\mathrm{max}}\hat{f}_c(x_i)$ to be in alignment with the accuracy of $\hat{f}$ on $x_i$ and points with confidences similar to that of $x_i$. As an example noted in \citet{guo2017calibration}, given a set of 100 predictions with confidences of $0.8$, we would hope that 80 of these predictions would be correctly classified. If so, we would consider the model to be \textit{calibrated}. Let $D^{test} = \{(x_i, y_i)\}_{i=1}^N$ be the test set on which $\hat{f}$ is evaluated, where each $x_i$ and $y_i$ are examples drawn from $(X, Y)$ (or a subset thereof), respectively. Further let $D^{test}_p = \{(x_i, y_i) \in D^{test} \hspace{1mm} \text{s.t.} \hspace{1mm} \hat{p}(x_i, \hat{f}) = p\}$. Formally, a model is calibrated if 

\begin{align}
    \label{eqn:ECE}
    \text{acc}(\hat{f}, D^{test}_p) = p\hspace{1mm}\forall\hspace{1mm}p \in [0, 1]
\end{align}

We further note as in \citet{guo2017calibration} that the probability in \ref{eqn:ECE} cannot be computed on a single sample, since an accuracy is computed on a set of examples rather than a single sample. Thus, there is a need for empirical approximations that can capture the essence of \ref{eqn:ECE}, which we describe below.

\begin{figure}
\centering
   \includegraphics[width=\textwidth]{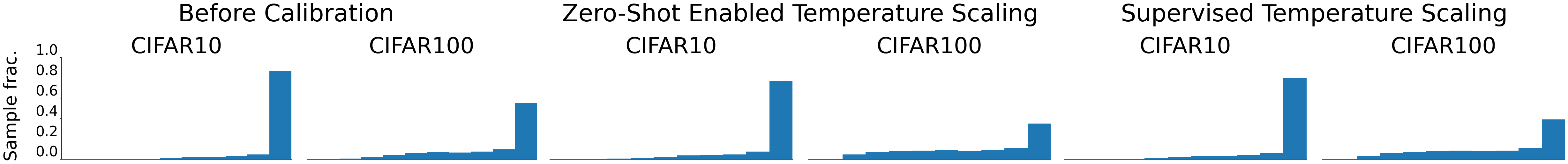}
   \includegraphics[width=\textwidth]{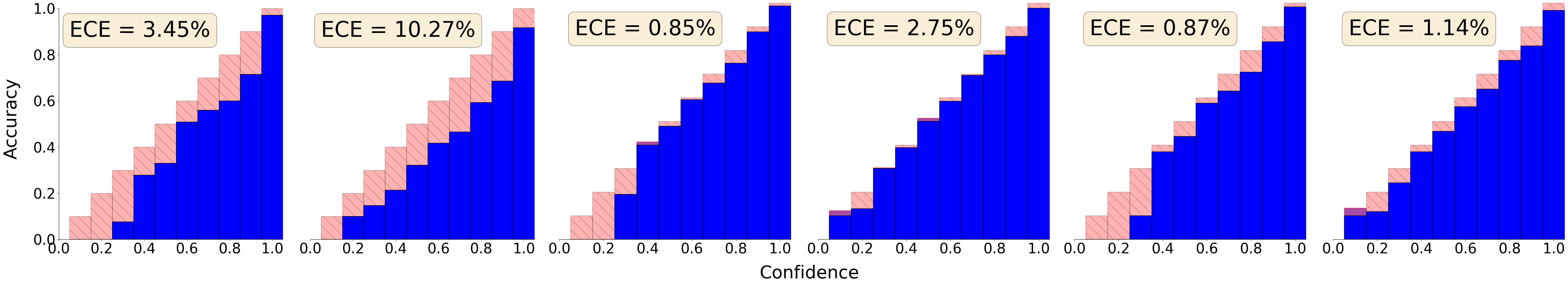}
\caption{Confidence histograms (top) and reliability diagrams (bottom) of \textbf{Left 2 Columns:} CLIP without any calibration, \textbf{Center 2 Columns:} CLIP with our method, Zero-Shot-Enabled Temperature Scaling, and \textbf{Right 2 Columns:} CLIP with Supervised Temperature Scaling. Miscalibration can be visualized by pink (overconfidence) and purple (underconfidence). Additionally, the Expected Calibration Error (ECE) for each evaluation is given, where a lower ECE is better. We note that our method is as calibrated and almost as calibrated as Supervised Temperature Scaling on CIFAR10 and CIFAR100, as measured by ECE, respectively, even though our method enables zero-shot inference. Here, we use ViT-B-16 \citep{dosovitskiy2020image} pretrained on laion400m\_e31 \citep{schuhmann2021laion} as our model, and ``a photo of \{\textit{CLASS NAME}\}" as our prompt.}
\label{fig:reliability_miscalibrated}
\end{figure}

\subsubsection{Visualizing Miscalibration Via Reliability Diagrams}

We visualize the calibration of our estimator through \textbf{reliability diagrams} \citep{degroot1983comparing, niculescu2005predicting}. These diagrams group points by their predicted confidence scores into $M$ equally spaced bins, and then compute the true and estimated accuracies in each bin as follows: let $B_m$ be the $(x_i, y_i)$ test samples whose confidence (i.e. estimated accuracy) falls into the interval $I_m = \big((m - 1)/M, m/M\big]$, for $m = 2, \dots, M$, and $I_1 = [0, \frac{1}{M}]$. The true accuracy is $\text{acc}(\hat{f}, B_m)$ and the \textit{estimated} accuracy (i.e. average confidence) within $B_m$ is $1/|B_m|\sum_{(x_i, y_i) \in B_m}\hat{p}(x_i, \hat{f})$, which we write in short-hand as $\hat{p}(\hat{f}, B_m)$. The reliability diagram plots the difference between true accuracy and estimated accuracy for all $M$ bins, and deviations from the line $f(x) = x$ represent miscalibrations: areas where there is a significant difference between the estimated and true accuracy. In Figure \ref{fig:reliability_miscalibrated}, the pink and purple portions of bars represent overconfidence and underconfidence, respectively, while the blue portions of bars represent how well-calibrated the model is. For all experiments, we let $M = 10$, as is standard \citep{guo2017calibration, rajendran2019accurate, kull2019beyond}.

Following standard practice \citep{minderer2021revisiting}, we also visualize a histogram of the number of points in each bin $|B_m|$. Mismatches between confidence and accuracy in bins with a relatively large amount of points are more grave than mismatches corresponding to bins with fewer points, since it means that the model was more miscalibrated on a larger number of points.

\subsubsection{Quantifying Miscalibration Via Expected Calibration Error (\textit{ECE})}

We can quantify this miscalibration with the \textbf{Expected Calibration Error (\textit{ECE})} introduced in \citet{10.5555/2888116.2888120}. \textit{ECE}, aimed at summarizing the miscalibration visualized in reliability diagrams, is calculated as 
 \begin{align}
     \text{ECE} = \sum_{m = 1}^M \dfrac{|B_m|}{|D|}\bigg\lvert\hat{p}(\hat{f}, B_m) - \text{acc}(\hat{f}, B_m) \bigg\rvert\label{eqn:miscalibration}
 \end{align}

In the leftmost columns of Figure \ref{fig:reliability_miscalibrated}, we show the calibration of CLIP, as it is regularly used, via reliability diagrams.

\subsection{Calibrating With Labels Via Temperature Scaling}
\label{calibrate_labels}
Typically used to reduce miscalibration, Temperature Scaling \citep{guo2017calibration} geometrically decreases the logit function $L$ by a scalar $T$. That is, $\hat{f}$ has a logit function that employs Temperature Scaling as $L^{\text{calibrated}}_c(x_i ; T) = L_c(x_i) / T$. In the case of CLIP, the Temperature-Scaling-infused logit function is $$L^{\text{calibrated}}_c(x_i ; T) = L^{\text{CLIP}}_c(x_i) / T$$Temperature Scaling typically calibrates a frozen network post-training using a dataset $D_{\text{calibration}} = \{(x_i, y_i)\}_{i=1}^N$ by minimizing the cross-entropy loss function $\mathcal{L}_{CE}(\hat{f}, D_{\text{calibration}}) = -\underset{(x_i, y_i) \in D_{\text{calibration}}}{\sum}\log\hat{f}_{y_i}(x_i)$.

In the rightmost two rows of Figure \ref{fig:reliability_miscalibrated}, we present reliability diagrams of CLIP once calibrated via Temperature Scaling. In Appendix Section \ref{appendix_section:supervised_methods}, we present reliability diagrams in this evaluation setting using additional supervised calibration methods Isotonic Regression introduced in \cite{10.1145/775047.775151}, 
%Platt Scaling introduced in \cite{article},
and Histogram Binning introduced in \cite{10.5555/645530.655658}, as well as Unsupervised Temperature Scaling introduced in \cite{mozafari2019unsupervised}. These are solely for context, since the main benefits of using CLIP are to be able to perform inference on a task of interest both without having any training labels on that task (which are required by the supervised methods) and without a calibration dataset that is specific for each inference dataset (which are required by all of the previously mentioned methods).

\begin{table}[t]
\centering
\begin{tabular}{ c c | c | c | c} 
  \hline
  \multicolumn{2}{c}{Method} & CLIP & CLIP + 0-Shot-Enabled TS & CLIP + TS \\
  \hline
    Architecture & Pre-Train Data & & & \\
  \cline{1-2}
  ViT-B-16 & laion400m & 6.34 & 2.22 & 0.91 \\
  & laion2b & 4.65 & 2.96 & 0.98 \\
   \hline
  ViT-L-14 & laion400m & 6.68 & 1.36 & 0.72 \\
  & laion2b & 3.17 & 2.38 & 0.85 \\
   \hline
  ViT-B-32 & laion400m & 4.69 & 3.06 & 1.66 \\
  &  laion2b & 3.88 & 2.69 & 0.80 \\
  \hline
  ViT-H-14 & laion2b & 3.67 & 2.47 & 0.88 \\
  \hline
  ResNet-50 & yfcc15m & 26.69& 7.60 & 2.61 \\
  & cc12m & 26.56 & 6.18 & 3.31 \\
  \hline

\end{tabular}
\caption{\textit{ECE} results on a variety of prompts, architectures, inference datasets, and pre-training datasets. Each results is the mean across 3 different inference datasets with different prompts for each dataset (per the original CLIP paper from \cite{minderer2021revisiting}) which are enumerated in Appendix Section \ref{appendix:prompt_dataset_enumeration}. For citations and links to the architectures and pre-training datasets, please see Appendix Section \ref{appendix:arch_pretrain_cite}. All numbers are percentages.}
\label{table:ECE_main}
\end{table}

\section{Our Method: Zero-Shot-Enabled Temperature Scaling}
To address this gap of the inability to calibrate CLIP without a calibration dataset, we propose Zero-Shot-Enabled Temperature Scaling. For a given architecture and pre-training dataset of CLIP, we simply train a temperature on an auxiliary dataset via Temperature Scaling. We then use this temperature on all downstream inferences of this model regardless of prompt or inference dataset. For all experiments, we use ImageNet-1k \citep{huang2021mos} as our auxiliary dataset with ``a photo of \{\}" as the prompt in the supervised training of the temperature ultimately used in our Zero-Shot-Enabled Temperature Scaling. Once trained, this model can be used for zero-shot inference since it does not require any re-training or tuning to be used on any given inference dataset with any given prompt. 

We note that the training of Zero-Shot-Enabled Temperature Scaling \textit{does} require a dataset on which to perform \textit{a} training process. However, this method enables zero-shot inference in an identical sense to CLIP: CLIP trains parameters on an auxiliary dataset and enables zero-shot inference on any arbitrary unseen distribution without any training on a dataset with significant distribution overlap to the inference distribution of interest. We note that CLIP allows the training of the parameters on a given architecture and a given pre-training dataset to be independent of the training of the parameters of different architectures and pre-training datasets. Therefore, we allow a different temperature for each architecture/pre-training dataset pair, matching the CLIP paradigm. Given this single parameter $T$ associated with the CLIP architecture and pre-training dataset, a user can utilize our method by simply diving the CLIP logits by $T$, without any training, tuning, or calibration processes necessary.

In the middle two columns of Figure \ref{fig:reliability_miscalibrated}, we present reliability diagrams of a CLIP-based model once updated via Zero-Shot-Enabled Temperature Scaling.

\section{Results}
\subsection{Comparison to Vanilla CLIP and CLIP + Temperature Scaling}
\begin{figure}[t]
\centering
   \includegraphics[width=\textwidth]{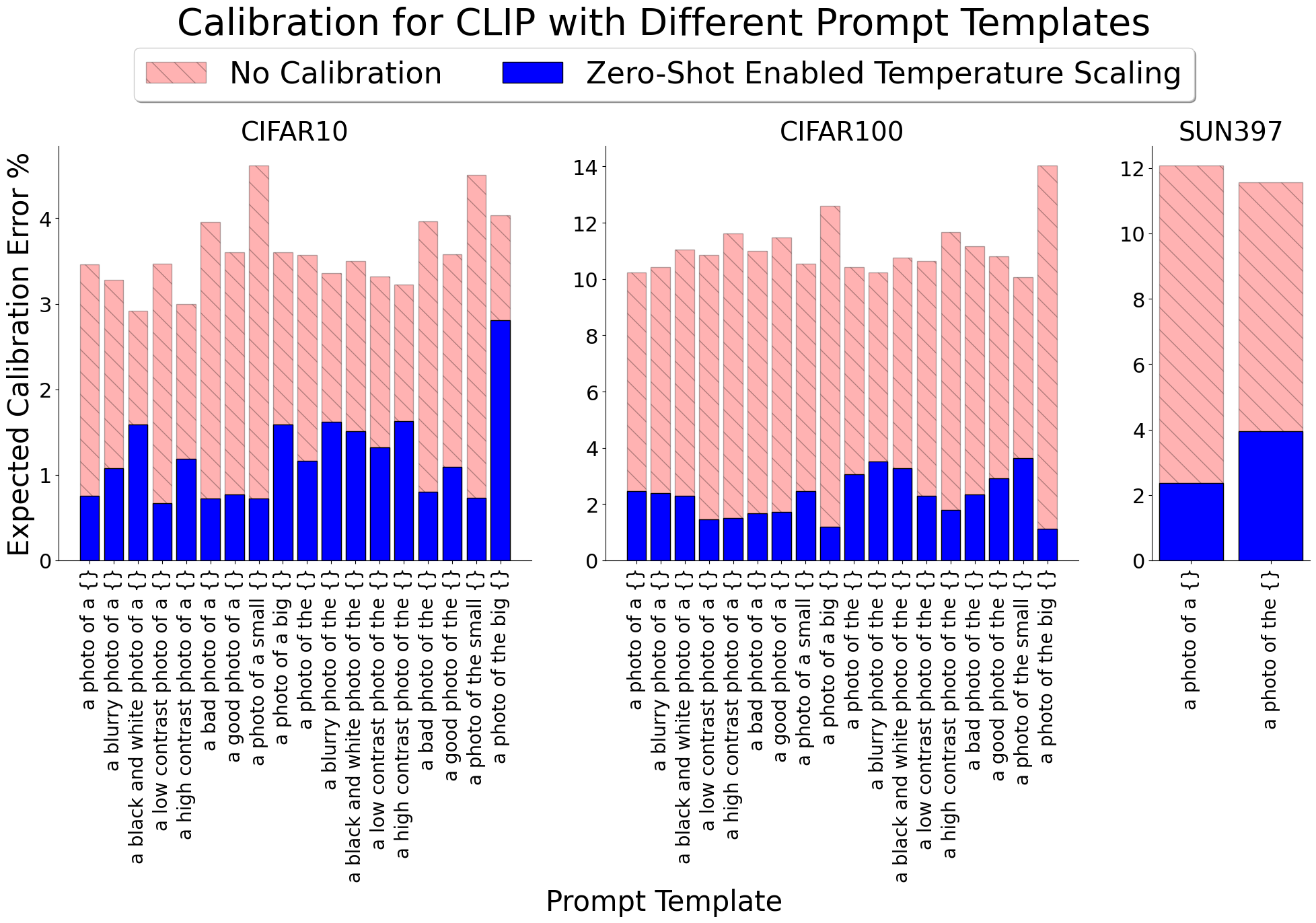}
\caption{Differences in \textit{ECE} with and without Zero-Shot-Enabled Temperature Scaling for various prompts and datasets. Regions where Zero-Shot-Enabled Temperature Scaling lowered \textit{ECE} (i.e. improved calibration) can be visualized in the pink portion of the bars. Here, we use ViT-B-16 (\cite{dosovitskiy2020image}) pretrained on laion400m\_e31 (\cite{schuhmann2021laion}) as our model.}
\label{fig:prompt_robustness}
\end{figure}

In Table \ref{table:ECE_main}, we show the Expected Calibration Error results of our method compared to CLIP and CLIP calibrated via supervised Temperature Scaling on a variety of prompts, architectures, inference datasets, and pre-training datasets. For detailed results across prompts and datasets for a single pre-training dataset and architecture (selected arbitrarily), please see Figure \ref{fig:prompt_robustness}. We note that our results are superior to CLIP without any calibration in all settings, but that our method results in models that are still significantly less calibrated than those calibrated via the supervised variant of Temperature Scaling - thus, leaving room for future Zero-Shot-Enabled CLIP calibration methods that improve upon our method.

\begin{figure}[t]
    \centering
    \includegraphics[width=.8\textwidth]{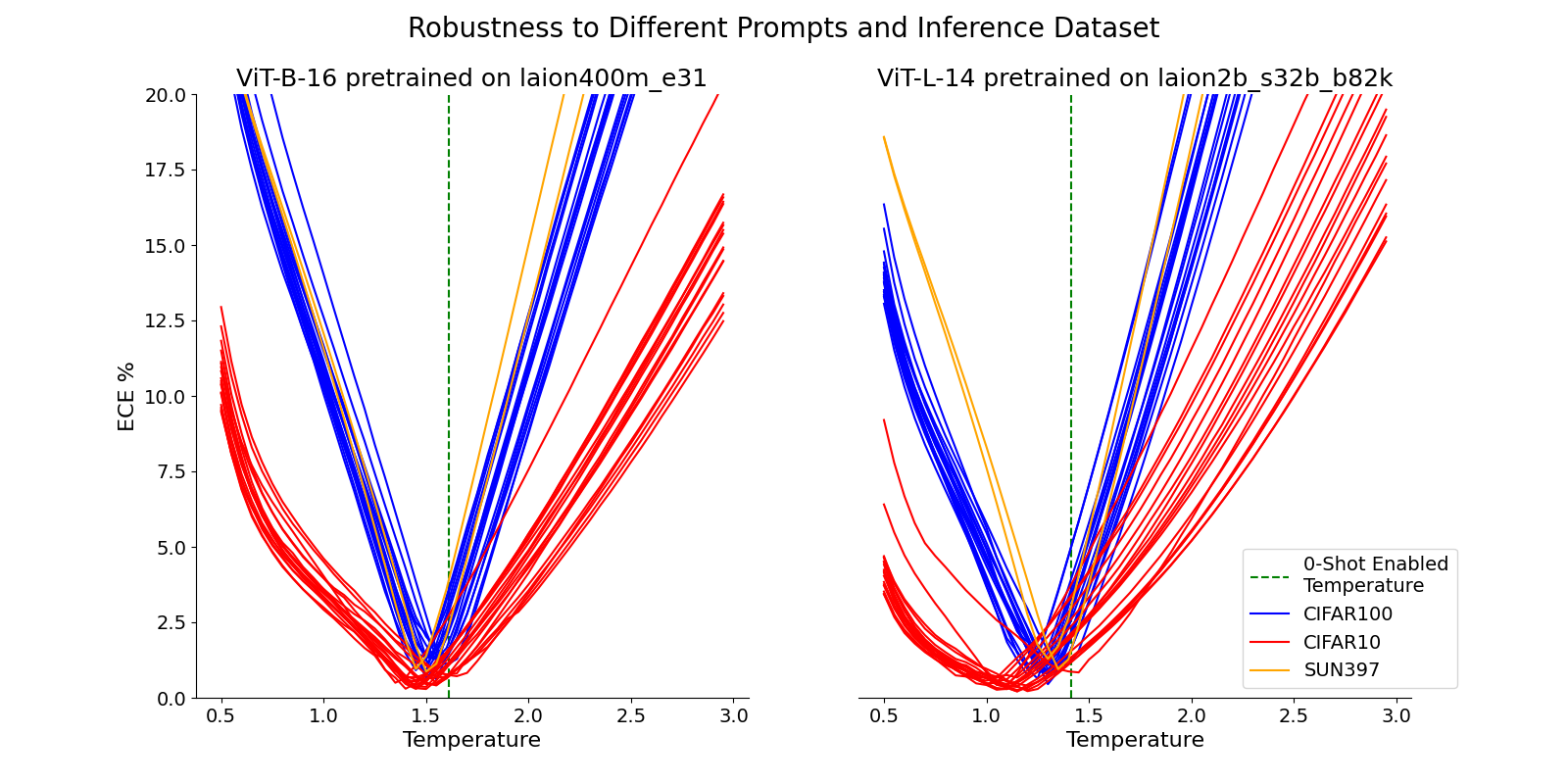}
    \caption{A plot of the effect of temperature on \textit{ECE} for various prompts and inference datasets (enumerated in Appendix Section \ref{appendix:prompt_dataset_enumeration}). Different lines of the same color represent different prompts. Note that the optimal temperature $T$ per inference dataset and prompt is very close to the one learned on a large auxiliary dataset and is approximately the same across inference datasets and prompts.}
    \label{fig:robust}
\end{figure}

\subsection{Robustness to Changes in Prompt and Inference Dataset}
In Figure \ref{fig:robust}, we show that, for a single architecture and pre-training dataset, the optimal temperature $T$ across prompts and inference datasets are approximately the same. This is remarkable considering each of the inference and auxiliary datasets has a different number of classes from each other (since CIFAR10 has 10 classes, CIFAR100 has 100 classes, SUN397 has 397 classes, and ImageNet-1k has 1000 classes), as well as different distributions. We do note that this temperature $T$ needs to be trained per architecute/pre-training dataset pair, as different pairs have (slightly) different optimal $T$'s, as can be visualized by comparing the approximate optimal $T$'s in the left plot (around $T = 1.55$) and in the right plot (around $T = 1.35$) of Figure \ref{fig:robust}.

\section{Conclusion and Future Work}

The miscalibration of supervised classification models has been extensively studied and improved via many calibration methods. Yet, prior to this paper, that has not been the case for vision-language models (like CLIP) with a different evaluation setup than traditional deep learning models. In this paper, we have shown that CLIP out-of-the-box is generally miscalibrated for a variety of experimental parameters. Lastly, to address this miscalibration, we have also presented a calibration method for CLIP that modifies inference with a single parameter that is aligned with the CLIP zero-shot-inference paradigm. Future work will extend additional supervised calibration methods to CLIP's zero-shot-inference setting and provide improvements to our method to close the gap in calibration between our Zero-Shot-Enabled Temperature Scaling and the supervised variant of Temperature Scaling.

\bibliography{iclr2023_conference}
\bibliographystyle{iclr2023_conference}

\appendix
\section{Appendix}
\subsection{Calibration Results of Supervised Methods}
\label{appendix_section:supervised_methods}

\begin{figure}[!htb]
\centering
   \includegraphics[width=\textwidth]{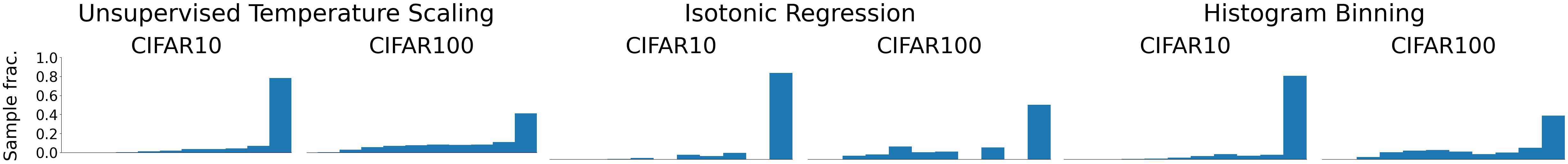}
   \includegraphics[width=\textwidth]{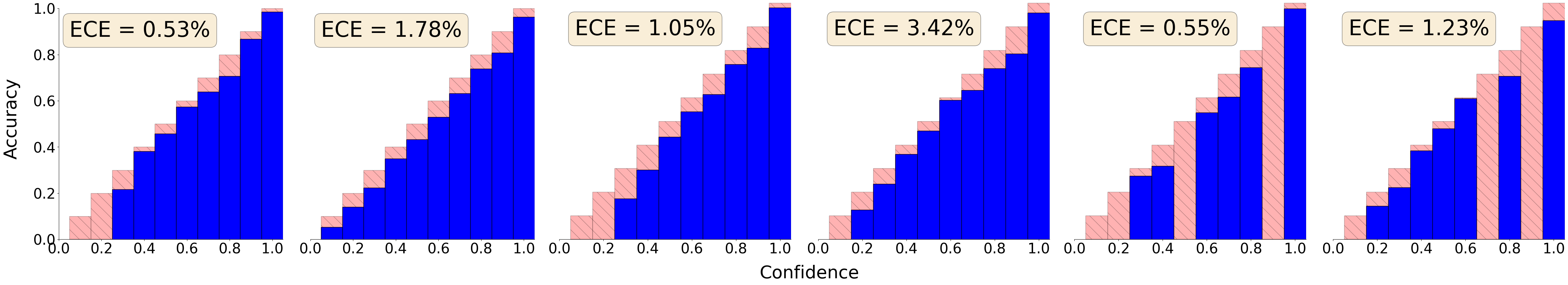}
\caption{Reliability diagrams of \textbf{Left 2 Columns:} CLIP with Unsupervised Temperature Scaling \citep{mozafari2019unsupervised}, \textbf{Center 2 Columns:} CLIP with Isotonic Regression \citep{10.1145/775047.775151}, and \textbf{Right 2 Columns:} CLIP with Histogram Binning \citep{10.5555/645530.655658}. Here, we use ViT-B-16 \citep{dosovitskiy2020image} pretrained on laion400m\_e31 \citep{schuhmann2021laion} as our model, and ``a photo of \{\textit{CLASS NAME}\}" as our prompt.}
\label{fig:supervised_reliability_miscalibrated}
\end{figure}

 In Figure \ref{fig:supervised_reliability_miscalibrated} we present reliability diagrams for CLIP using the following supervised calibration methods: Isotonic Regression \citep{10.1145/775047.775151} and Histogram Binning \citep{10.5555/645530.655658}, as well as Unsupervised Temperature Scaling \citep{mozafari2019unsupervised}. All three of these methods perform well and significantly reduce miscalibration. However, as mentioned in Section \ref{calibrate_labels}, the use of these methods is inconsistent with how CLIP is often used, and is therefore impractical for wide-scale adoption.
 
\subsection{Prompts and Datasets Used in Figure \ref{table:ECE_main}}
\label{appendix:prompt_dataset_enumeration}
Our \textit{ECE} results in Figure \ref{table:ECE_main} are averaged over the following datasets and prompts:
\begin{enumerate}
        \item SUN397 \citep{yu2015lsun} with the following prompts:
            \begin{enumerate}
                \item ``a photo of \{\}"
                \item ``a photo of the \{\}"
            \end{enumerate}
        \item CIFAR10 and CIFAR100 \citep{Krizhevsky09learningmultiple} with the following prompts:
            \begin{enumerate}
                \item ``a photo of a \{\}"
                \item ``a blurry photo of a \{\}"
                \item ``a black and white photo of a \{\}"
                \item ``a low contrast photo of a \{\}"
                \item ``a high contrast photo of a \{\}"
                \item ``a bad photo of a \{\}"
                \item ``a good photo of a \{\}"
                \item ``a photo of a small \{\}"
                \item ``a photo of a big \{\}"
                \item ``a photo of the \{\}"
                \item ``a blurry photo of the \{\}"
                \item ``a black and white photo of the \{\}"
                \item ``a low contrast photo of the \{\}"
                \item ``a high contrast photo of the \{\}"
                \item ``a bad photo of the \{\}"
                \item ``a good photo of the \{\}"
                \item ``a photo of the small \{\}"
                \item ``a photo of the big \{\}"
            \end{enumerate}
    \end{enumerate}

\subsection{Architectures and Pretraining Datasets Used in Figure \ref{table:ECE_main}}
\label{appendix:arch_pretrain_cite}

All models in Figure \ref{table:ECE_main} are from OpenCLIP \citep{ilharco_gabriel_2021_5143773}. All ViT models used are pretrained on either LAION-400M \citep{schuhmann2021laion} or LAION-2B \citep{schuhmann2022laionb}. The ResNet models are pretrained on YFCC15M, a subset of YFCC100M \citep{Thomee_2016}, or the Conceptual Captions Dataset \citep{changpinyo2021conceptual}
\end{document}